\documentclass{article} 
\usepackage{tencent_ailab_tech_report}
\usepackage[colorlinks = true,
            linkcolor = blue,
            urlcolor  = blue,
            citecolor = blue,
            anchorcolor = blue]{hyperref}

\usepackage{graphicx}
\usepackage{caption}
\usepackage{subcaption}
\usepackage{booktabs} 

\usepackage{listings}
\usepackage{xcolor}

\lstdefinelanguage{json}{
    basicstyle=\ttfamily\small,
    numbers=left,
    numberstyle=\tiny,
    stepnumber=1,
    numbersep=5pt,
    showstringspaces=false,
    breaklines=true,
    frame=lines,
    backgroundcolor=\color{gray!10},
    literate=
     *{0}{{{\color{blue}0}}}{1}
      {1}{{{\color{blue}1}}}{1}
      {2}{{{\color{blue}2}}}{1}
      {3}{{{\color{blue}3}}}{1}
      {4}{{{\color{blue}4}}}{1}
      {5}{{{\color{blue}5}}}{1}
      {6}{{{\color{blue}6}}}{1}
      {7}{{{\color{blue}7}}}{1}
      {8}{{{\color{blue}8}}}{1}
      {9}{{{\color{blue}9}}}{1}
      {:}{{{\color{red}:}}}{1}
      {,}{{{\color{red},}}}{1}
      {"}{{{\color{orange}"}}}{1}
      {[}{{{\color{green}[}}}{1}
      {]}{{{\color{green}]}}}{1}
      {\{}{{{\color{green}\{}}}{1}
      {\}}{{{\color{green}\}}}}{1}
}

\usepackage{microtype}
\usepackage{hyperref}
\usepackage{url}
\usepackage{booktabs}
\usepackage{enumitem}
\usepackage{multicol}
\usepackage{multirow}
\usepackage{CJKutf8}
\usepackage{amsmath}
\usepackage{siunitx}
\usepackage{floatflt}
\usepackage{wrapfig}
\usepackage{graphicx}
\usepackage{booktabs}
\usepackage{wrapfig}
\usepackage{authblk}
\usepackage{lipsum}
\usepackage{dsfont}
\usepackage{bbm}
\usepackage{algorithm}
\usepackage{algorithmicx}
\usepackage{algpseudocode}
\usepackage{microtype}
\usepackage{graphicx}
\usepackage{subcaption}
\usepackage{multirow}
\usepackage{booktabs} 
\usepackage{pifont}  
\usepackage{graphicx}  
\usepackage{subcaption} 
\usepackage{hyperref}
\usepackage[most]{tcolorbox}
\newtcolorbox{alprompt}[1]{
        boxrule = 1pt,
        fontupper = \small\tt,
        fonttitle = \bf\color{black},
        arc = 2pt,
        rounded corners,
        colframe = black,
        colbacktitle = white!97!yellow,
        colback = white!97!yellow,
        title = #1,
}
\definecolor{darkgreen}{rgb}{0.0, 0.5, 0.0}
\definecolor{darkgray}{gray}{0.4}
\definecolor{maroon}{rgb}{0.5, 0.0, 0.0}
\definecolor{navy}{rgb}{0.0, 0.0, 0.5}
\definecolor{teal}{rgb}{0.0, 0.5, 0.5}

\definecolor{deepblue}{RGB}{41, 128, 185}

\definecolor{mylightgreen}{RGB}{144,238,144}
\definecolor{mylightblue}{RGB}{173,216,230}

\definecolor{outerboxcolor}{gray}{0.90} 
\definecolor{innerboxcolor}{rgb}{1,1,1}

\definecolor{nred}{RGB}{196, 38, 11}
\definecolor{ngreen}{RGB}{18, 141, 21}
\definecolor{nblue}{RGB}{41, 52, 190}

\usepackage{listings}
\usepackage{xcolor}

\algnewcommand{\LeftComment}[1]{\Statex \(\triangleright\) #1}

\usepackage{array}
\usepackage{amsmath}
\usepackage{mathtools}
\usepackage{amsthm}
\usepackage{arydshln}
\usepackage[capitalize,noabbrev]{cleveref}
\usepackage{adjustbox} 
\usepackage{enumitem}
\usepackage{longtable}

\theoremstyle{plain}

\theoremstyle{definition}

\theoremstyle{remark}

\usepackage[textsize=tiny]{todonotes}

\makeatletter
\renewcommand{\@fnsymbol}[1]{\ensuremath{\ifcase#1\or \dag\or \ddag\or \S\or \P\or \|\or **\fi}}
\makeatother

\newcommand\blfootnote[1]{%
  \begingroup
    \renewcommand\thefootnote{}\footnote{#1}%
  \addtocounter{footnote}{-1}%
  \endgroup
}

\colmfinalcopy

\title{MobileGUI-RL: Advancing Mobile GUI Agent through Reinforcement Learning in Online Environment}

\author[ ]{Yucheng Shi$^{1,2 *}$, Wenhao Yu$^{1 *}$, Zaitang Li$^{1,3}$, Yonglin  Wang$^{1}$, Hongming Zhang$^{1}$, Ninghao Liu$^{2}$, Haitao Mi$^{1}$, Dong Yu$^{1}$}

\affil[ ]{$^{1}$Tencent AI Seattle Lab, $^{2}$University of Georgia, $^{3}$Chinese University of Hong Kong}

\begin{document}

\maketitle

\blfootnote{* equal contribution}

\begin{abstract}
Recently, there has been a surge of vision-based GUI agents designed to automate everyday mobile and web tasks. These agents interpret raw GUI screenshots and autonomously decide where to click, scroll, or type, which bypasses handcrafted rules and app-specific APIs.
However, most existing methods trained GUI agent in the offline environment using pre-collected trajectories. 
This approach limits scalability, causes overfitting to specific UI templates, and leads to brittle policies when faced with unseen environment. 
We present \textit{MobileGUI-RL}, a scalable framework that trains GUI agent in online environment. MobileGUI-RL contains two key components. It (i) synthesizes a curriculum of learnable tasks through self-exploration and filtering, and (ii) adapts GRPO to GUI navigation with trajectory-aware advantages and composite rewards that balance task success and execution efficiency. 
Experiments on three online mobile-agent benchmarks show consistent gains, validating the effectiveness of our approach.
\end{abstract}

\section{Introduction}

Recent advances in large vision-language models (LVLMs)~\citep{hurst2024gpt,anthropic2025claude,bai2025qwen2} have opened up new possibilities for building vision-based GUI agents~\citep{qin2025ui,xu2025aguvis}, fundamentally transforming the way intelligent agents interact with graphical user interfaces (GUIs). 
Unlike traditional pipeline-based GUI agents, which typically decompose the task into separate planning and GUI grounding stages~\citep{zheng2024gpt,gou2025navigating}, these vision-based GUI agents leverage the powerful perception and reasoning abilities of LVLMs to directly interpret GUI screenshots and autonomously determine actions such as clicking, scrolling, and typing~\citep{wang2024gui,zhang2024large}.
By eliminating the need for handcrafted rules or access to underlying application APIs, vision-based GUI agents offer a flexible, scalable, and platform-agnostic solution for automating interactions across a wide range of apps and devices.

\begin{figure}[t]
    \centering
    \vspace{-0.2in}
    \includegraphics[width=1\linewidth]{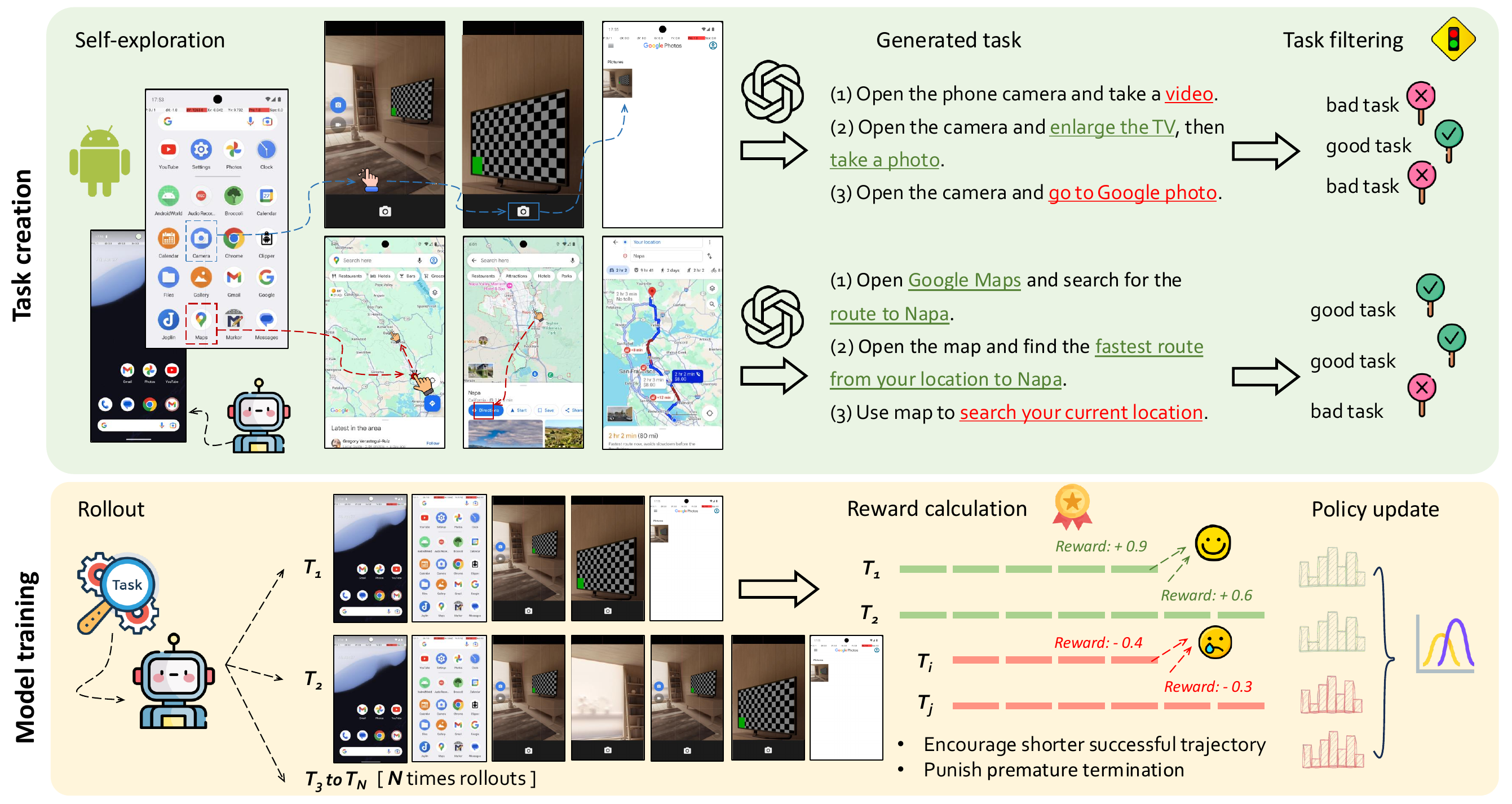}
    \vspace{-0.3in}
    \caption{Framework overall -- a scalable pipeline for training GUI agents through self-exploration, task filtering, and trajectory-level reinforcement learning with a structured reward design.}
    \vspace{-0.1in}
    \label{fig:enter-label}
\end{figure}

Despite these advances, training GUI agents that can operate robustly in real-world environments remains a highly challenging task. Most existing method train GUI agents in the offline environment that rely on static, pre-collected trajectory data for supervised fine-tuning~\citep{wu2024atlas,qin2025ui,sun2025genesis}.
Another line of research explores step-wise reinforcement learning, inspired by the recent DeepSeek-R1 paradigm~\citep{zhou2025gui,lu2025ui,luo2025gui}.
However, offline learning methods rely extensively on high-quality annotations for action trajectories, which require step-by-step executions and precise evaluations of their correctness. 
Such detailed annotations are labor-intensive and challenging to scale~\citep{wu2024atlas,qin2025ui}.
Moreover, GUI agents trained with SFT or offline reinforcement learning often overfit to specific interface patterns~\citep{qin2025ui,sun2025genesis,xu2025aguvis}.
Such overfitting leads to poor generalization when encountering task instructions deviating from familiar templates or to dynamic UI environments. In practice, real-world GUIs are highly variable: new screens frequently emerge, interface elements change or disappear unpredictably, and user interactions can substantially alter GUI states.
Pre-trained policies struggle to adapt to such changes, limiting real-world usability.

To address these limitations, training GUI agents in online environments has emerged as a promising direction~\citep{bai2024digirl,wang2025distrl}, enabling agents to continuously interact with their environment and update policies in real time. However, it introduces several challenges.
First, online learning requires real-time interaction with the environment at every training step. Each action must be executed, and its effect observed, before updating the policy. This process can be slow and computationally expensive, especially when scaling to complex apps or mobile devices where GUI rendering and response times vary.
Second, defining meaningful trajectory-level reward signals is nontrivial. Many tasks have long trajectories, where the agent must execute a sequence of steps before achieving a goal. At the same time, multiple action sequences may lead to the same outcome, and near-correct trajectories can fail due to a single misstep. These challenges make reward-driven learning difficult, potentially slowing convergence and leading to suboptimal policies.

In this work, we present MobileGUI-RL, a novel framework for training GUI agents through reinforcement learning in online environments. To support this, we develop an interactive environment that supports virtual machine management and continuous online learning, enabling agents to explore and adapt to the full spectrum of mobile GUI interactions.
MobileGUI-RL consists of two key components. First, we employ a synthetic task generation pipeline that combines self-exploration with filtering, producing a curriculum of learnable tasks tailored to the agent's current capabilities.
Then, we adapt group relative policy optimization (GRPO)~\citep{shao2024deepseekmath,guo2025deepseek} and introduce a trajectory-aware advantage and multi-component rewards that balance task success and execution efficiency.
Experiments on four mobile agent benchmarks show that MobileGUI-RL improves performance in both online and offline evaluations. We also observe a steady improvement in online performance throughout the reinforcement learning process.

\section{Related Work}

\subsection{GUI Agent}
The paradigm for autonomous GUI interaction has recently shifted towards agents powered by large vision language models (LVLMs) as their core reasoning engine~\citep{gur2023real,zhang2024large,wang2024gui}. These agents typically interpret visual data from screenshots, sometimes augmented with structural information, to predict their actions. Earlier approaches often employed multi-stage pipelines with separate planning and grounding phases~\cite{zheng2024gpt,gou2025navigating}. However, the field is steadily shifting toward end-to-end models that operate directly on raw pixels, offering a more scalable and human-like framework for UI interaction~\citep{hong2024cogagent,xu2025aguvis,qin2025ui}. 
While recent methods have enhanced agent reasoning, navigating complex and dynamic mobile environments still demands more advanced planning and adaptability. For example, Aguvis~\citep{xu2025aguvis} augments datasets with language model generated chain-of-thought annotations, while UI-TARS~\citep{qin2025ui} incorporates both positive and negative examples to support self-reflection and error correction via direct preference optimization. Other approaches improve navigation and error recovery by giving agents explicit control over their trajectories—such as the ability to rollback to prior states or explore alternative action paths~\citep{zhang2025enhancing, hu2025webcot}. To address stagnation in self-improving agents, WebEvolver~\citep{fang2025webevolver} introduces co-evolving world models for look-ahead simulation. In contrast to prior work that relies on offline tuning or static environments, MobileGUI-RL adopts online reinforcement learning from live interaction trajectories, enabling real-time policy updates and more adaptive performance.




\subsection{RL with Agent}
Recent advancements in GUI agents have marked a significant shift from reliance on supervised fine-tuning (SFT) to the adoption of reinforcement learning (RL) to improve generalization and decision-making capabilities. This trend is largely inspired by the success of DeepSeek-R1~\citep{guo2025deepseek}. The ``R1-style'' training paradigm, which uses RL to directly optimize policies based on task rewards, has been effectively applied to the agent domain. For example, GUI-R1 demonstrated that RL can achieve state-of-the-art performance on GUI tasks across multiple platforms while using only a fraction of the data required by SFT methods~\citep{luo2025gui}. Similarly, WebAgent-R1 employed a multi-turn RL framework for web navigation, significantly boosting base model success rates by learning directly from online interactions with binary success signals~\citep{wei2025webagent}.

While the direct application of RL has proven effective, the unique challenges of long-horizon interaction, sparse rewards, and high data cost inherent to GUI and web agent tasks have spurred the development of more specialized algorithms. To improve credit assignment in multi-step tasks, GiGPO improves credit assignment with hierarchical advantage estimation~\citep{feng2025group}, while ARPO enhances GRPO with replay buffers for better sample efficiency~\citep{lu2025arpo}. To reduce reliance on human-annotated data, DigiRL adopts an autonomous offline-to-online RL pipeline~\citep{bai2024digirl}. Complementary efforts like ZeroGUI use VLMs for automatic task and reward generation~\citep{yang2025zerogui}, and InfiGUI-R1 evolves agents from reactive to deliberative through RL-based planning and recovery~\citep{liu2025infigui}.

\section{MobileGUI-RL}

\subsection{Overview}
We formulate the problem of GUI task completion as a Markov Decision Process (MDP), defined by the tuple $\mathcal{M} = (\mathcal{S}, \mathcal{A}, \mathcal{P}, \mathcal{R})$~\citep{fang2025webevolver, hu2025webcot}. Here, $\mathcal{S}$ represents the state space of GUI screenshots and system states, $\mathcal{A}$ encompasses the action space of user interactions (e.g., taps, swipes, text input), $\mathcal{P}$ denotes the transition determined by the mobile operating system, $\mathcal{R}$ is a reward function that evaluate task completion. 
Given a natural language instruction $\mathbf{q}$, our goal is to train an agent to learn a policy $\pi_\theta(\mathcal{A} \mid \mathcal{S}, \mathbf{q})$ to complete the given task accurately while maximizing the expected cumulative reward over time. This MDP formulation provides a principled framework for learning and evaluating interactive agents in complex, dynamic mobile GUI environments.


To train GUI agents using online trajectory reinforcement learning, we propose three novel modules, detailed as follows:
First, we design an interactive environment that supports continuous online learning, enabling agents to explore and adapt across the full spectrum of mobile GUI interactions (Section ~\ref{sec:method.1}). Second, we introduce a synthetic task generation pipeline that combines self-exploration with task filtering, yielding a dynamic curriculum tailored to the agent’s evolving capabilities (Section~\ref{sec:method.2}). Third, we adapt Group Relative Policy Optimization (GRPO) to the unique challenges of GUI navigation, incorporating trajectory-aware advantage estimation and a multi-component reward structure that balances task success with execution efficiency (Section~\ref{sec:method.3}).

\subsection{Scalable and Interactable Environment for Online Learning}
\label{sec:method.1}

To support GUI agents with online trajectory reinforcement learning, we design a training environment centered on two key capabilities: \textit{batched virtual execution} and \textit{real-time agent interaction}.


\textbf{Batched Virtual Execution.} 
At the core of our system is a scalable, asynchronous framework that deploys multiple Android emulator~\citep{android_emulator} instances in parallel across CPU machines. This batched execution enables agents to interact with diverse GUI environments simultaneously, significantly increasing throughput and trajectory diversity. Trajectories are collected asynchronously from a pool of emulators running in parallel, while policy optimization is performed separately on GPU servers. This architecture aligns compute-intensive environment simulation with CPU resources and model training with GPUs, optimizing overall resource utilization. 
Moreover, the asynchronous design improves scalability, allowing more environment instances to be added without bottlenecking training, even when hardware performance varies.
As a result, the framework supports large-scale rollout and yields more diverse interaction trajectories, enhancing the robustness and generalization of the learned policies across real-world mobile GUI tasks.

\textbf{Real-Time Agent Interaction.} 
At each timestep $t$, the agent observes a multimodal state representation $s_t = (v_t, \mathbf{q}, \mathbf{h}_t)$ comprising three essential components~\citep{zheng2024gpt}. The visual input $v_t$ provides the current screenshot capturing the complete GUI state. The task goal $\mathbf{q}$ specifies the natural language instruction that guides the agent's behavior. The interaction history $\mathbf{h}_t = \{(s_0, a_0), ..., (s_{t-1}, a_{t-1})\}$ maintains temporal context, enabling the agent to reason about past actions and their consequences. More details on input construction are in appendix \textbf{X}.

The agent, a vision-language model in our setting, will process this state representation and first generate an internal reasoning trace $\mathbf{c}_t$, then produce a structured action $a_t \in \mathcal{A}$. Our action space comprehensively covers mobile interactions through four categories: (1) Physical gestures include parameterized actions such as $\text{tap}(x, y)$ and $\text{swipe}(x_1, y_1, x_2, y_2)$ using normalized coordinates $\in [0, 1]^2$ for resolution independence; (2) Text input actions $\text{type}(\text{string})$ handle keyboard interactions; (3) System navigation encompasses device-level operations including $\{\text{back}, \text{home}, \text{recent}\}$; (4) Control actions include $\text{wait}(t)$ for synchronization with dynamic UI elements and $\text{terminate}(\text{status})$ for episode completion. More detailed action definitions are provided in the appendix section.

A sequence of these interactions $\tau = (s_0, a_0, s_1, a_1, ..., s_T)$ forms a \textit{trajectory}, which is evaluated upon termination. 
Rather than relying on hand-crafted reward functions that poorly generalize across tasks, we employ a powerful vision-language model oracle $\mathcal{O}$ (e.g., Qwen 2.5 VL 72B) to serve as a unified evaluator. Given the final $k$ screenshots of a trajectory and the initial instruction $\mathbf{q}$, the oracle analyzes whether the task has been completed as intended: $r = \mathcal{O}(\{s_{T-k+1}, ..., s_T\}, \mathbf{q})$~\citep{bai2024digirl}. This evaluation produces a binary success signal that abstracts away low-level UI details, enabling scalable supervision across diverse GUI tasks without task-specific engineering.

\subsection{Synthetic Task Generation and Filtering}
\label{sec:method.2}

A critical challenge in training GUI agent in online environment is obtaining a diverse yet learnable curriculum of tasks. Real-world task distributions are heavily skewed toward common interactions, limiting the agent's ability to handle edge cases. Moreover, manually curating tasks is labor-intensive and fails to scale with the complexity of modern mobile ecosystems. We address this through a two-stage pipeline that automatically generates and filters synthetic tasks.

\subsubsection{Self-Exploration for Diverse Task Discovery}

Our self-exploration mechanism leverages the natural structure of mobile interfaces to discover meaningful tasks. The process begins with an exploration agent $\pi_{\text{explore}}$ performing random walks through the GUI environment. These walks are not purely random but incorporate basic heuristics such as preferring unexplored UI elements and avoiding repetitive loops. Each exploration trajectory $\tau_{\text{explore}} = \{(s_0, a_0), ..., (s_n, a_n)\}$ captures a sequence of state transitions that potentially represent a coherent task.
Inspired by \cite{sun2025genesis}, we then employ GPT-4o to reverse-engineer task descriptions from these trajectories. Given a trajectory, the model generates a natural language instruction $\mathbf{q}$ that would motivate the observed sequence of actions. This reverse process -- figuring out the goal from the actions -- produces a variety of tasks that match what the app is designed to do. The generated tasks span a wide spectrum, from simple interactions ("Open the settings menu") to complex multi-step procedures ("Set a recurring alarm for weekdays at 7 AM").

\subsubsection{Task Filtering via Text-based World Model}
While self-exploration can generate a wide range of task instructions, many of them suffer from two key issues: they are either too ambiguous, often due to limitations in the reverse-engineered LLM’s summarization ability, or too complex to be solved given the current GUI state and context. Attempting to execute these infeasible tasks leads to wasted computational effort and the generation of low-quality trajectories that may destabilize learning. To address this, we propose a lightweight filtering mechanism based on a text-based world model, which pre-screens candidate tasks before rollout. This approach effectively avoids unnecessary environment interactions while ensuring that the selected tasks are within the agent’s current capability.

Our filter first employs a LLM as a simulator $\mathcal{W}$ that can generate a textual representation $\tilde{s}$ of the GUI state. Given a task $\mathbf{q}$ and current state description $\tilde{s}$, the world model predicts the next state $\tilde{s}' = \mathcal{W}(\tilde{s}, a, \mathbf{q})$ resulting from action $a$. The filtering process proceeds as follows:
The world model first initializes with a textual description of the home screen, structured as a list of UI elements with their properties: $\tilde{s}_0 = \{e_1: \text{(type, content, bounds)}, ..., e_n: \text{(type, content, bounds)}\}$. Our base agent $\pi_{\text{base}}$ receives both the task and this description and outputs an action. The world model simulates the action's effect, generating a new state description that reflects the expected GUI changes. This process continues until the base agent signals task completion, failure, or exceeds a step limit $T_{\text{max}}$.

A task is admitted to the training set only if the simulation reaches a success state within the step limit: $\mathcal{F}(\mathbf{q}) = \mathbb{1}[\exists t \leq T_{\text{max}}: \pi_{\text{proxy}}(a_t|\tilde{s}_t, \mathbf{q}) = \text{terminate(success)}]$. This filtering process serves not only to remove logically inconsistent or overly complex tasks, but also plays a key role in decoupling perception from reasoning. Since the world model operates entirely on structured textual representations of the GUI, it removes the need for low-level visual grounding. As a result, the evaluation focuses solely on whether the agent’s reasoning and planning abilities are sufficient to solve the task, assuming perfect perception. This abstraction allows us to efficiently assess task feasibility and construct a curriculum without being confused by perception errors.


\subsection{Online Learning with MobGRPO}
\label{sec:method.3}

Training GUI agent in online environments presents several unique challenges. Rewards are typically sparse, trajectories can be long with variable step counts, and task outcomes often depend on delayed success signals. These challenges make credit assignment difficult and destabilize training under standard policy gradient methods such as PPO~\citep{schulman2017proximal}. To address these issues, we extend GRPO~\citep{shao2024deepseekmath} with a trajectory-aware formulation and a carefully designed reward structure for mobile GUI agent training, forming our proposed \textit{MobGRPO} algorithm.

\subsubsection{Trajectory-Aware Policy Optimization}

Our MobGRPO objective builds upon GRPO to handle variable-length trajectories and fine-grained action steps. For a batch of $G$ trajectories $\{\tau_i\}_{i=1}^G$ generated for task $\mathbf{q}$, we define the loss as:

\begin{equation}
\mathcal{L}_{\text{MobGRPO}} = -\frac{1}{\sum_{t=1}^{G}|\mathbf{o}_{i,s}|} \sum_{i=1}^{G} \sum_{s=1}^{S_i}  \sum_{t=1}^{|\mathbf{o}_{i,s}|} 
\left\{
\min \left[
r_t(\theta)\hat{A}_{i,s,t}, \,
\text{clip}\left(
r_t(\theta),
1 - \epsilon, 1 + \epsilon
\right) \hat{A}_{i,s,t}
\right]
\right\}
\end{equation}

where $r_t(\theta) = \frac{\pi_\theta(o_{s,t}|s_{<s}, a_{<s}, o_{s,<t})}{\pi_{\theta_{\text{old}}}(o_{s,t}|s_{<s}, a_{<s}, o_{s,<t})}$ is the token-level probability ratio in the action sequence, and $\hat{A}_{i,s,t}$ shares a trajectory-level advantage signal.
Instead of computing per-step rewards and advantages, we evaluate the entire trajectory $\tau$ after completion to obtain a single scalar reward $R(\tau, \mathbf{q})$ that reflects its overall quality. This trajectory-level reward is then used to compute a normalized advantage:
$
\hat{A}_{\tau} = \frac{R(\tau, \mathbf{q}) - \bar{R}_{\mathbf{q}}}{\sigma_{R_{\mathbf{q}}} + \epsilon},
$
where $\bar{R}_{\mathbf{q}}$ and $\sigma_{R_{\mathbf{q}}}$ denote the mean and standard deviation of trajectory rewards for task $\mathbf{q}$. This advantage is uniformly assigned to all steps within the trajectory, providing a consistent learning signal regardless of the trajectory length or where the success occurs.

By aggregating the reward at the trajectory level and distributing the advantage across all steps, our approach avoids noisy or misleading per-step supervision and addresses the credit assignment problem in long-horizon GUI tasks.

\subsubsection{Multi-Component Reward Design}

Reward design plays a central role in learning effective policies for GUI navigation, where tasks are long-horizon, rewards are sparse, and outcomes are often binary. Standard reward functions, e.g., assigning $r=1$ for success and $r=0$ otherwise, are inadequate in this context. They fail to differentiate between successful trajectories of varying quality and offer no learning signal when all rollouts in a batch succeed or fail uniformly. To address these limitations, we propose a multi-component reward function that captures trajectory-level quality, discourages premature termination, and ensures continuous learning signals for policy optimization.

\noindent\textbf{Differentiating Successful Trajectories.}  
Although many trajectories may successfully complete a task, they can differ significantly in terms of efficiency. In GUI settings, shorter trajectories are generally preferred as they reduce user friction and lower the risk of compounding errors. To reflect this, we introduce an exponentially decaying efficiency factor that rewards faster completions more,

\begin{equation}
f_{\text{efficiency}}(|\tau|) = \text{clip}(e^{-\lambda|\tau|}, \alpha_{\min}, \alpha_{\max}).
\end{equation}

This design not only encourages efficient behavior but also addresses a key issue in GRPO-like methods: when all trajectories succeed and receive identical rewards (e.g., $r=1$), the normalized advantage becomes zero, halting policy updates. Our reward structure introduces relative differences even among successful rollouts, preserving gradient signals for continued learning.

\noindent\textbf{Penalizing Premature Termination.}  
Agents trained on sparse-reward environments often learn to "give up" early when facing difficult or ambiguous tasks, terminating episodes prematurely before fully attempting or exploring the instruction. To discourage this behavior, we introduce a penalty for early exits when the task is not yet completed:

\begin{equation}
g(|\tau|) = 1 - \frac{|\tau|}{T_{\max}}, \quad \text{penalty} = \beta_{\max} \cdot g(|\tau|)
\end{equation}

This linear decay penalizes early termination more heavily than later exits, encouraging the agent to engage more thoughtfully with the task before deciding to stop.

\noindent\textbf{Handling Degenerate Batches.}  
Another practical issue arises when all trajectories in a batch fail, yielding zero rewards. In such cases, the computed advantages are uniformly zero, resulting in no policy update. We adopt the same mitigation as proposed in DAPO~\citep{yu2025dapo} by filtering out these degenerate batches during training to maintain meaningful optimization dynamics.

\noindent\textbf{Final Reward Formulation.}  
Combining these components, our composite reward is defined as:

\begin{equation}
R(\tau, \mathbf{q}) = \begin{cases}
r_{\text{base}} \cdot f_{\text{efficiency}}(|\tau|) & \text{if success} \\
-\beta_{\max} \cdot g(|\tau|) & \text{if fail}
\end{cases}
\end{equation}

This formulation delivers a dense, interpretable, and differentiable learning signal that encourages success, promotes efficiency, penalizes shortcuts, and maintains update dynamics across varying batch conditions. The modular design also allows fine-tuning through hyperparameters to suit deployment-specific requirements.

\section{Experiments}
\subsection{Experiments Setting}
\subsubsection{Parameter Settings}
Our training environment is built on a scalable pool of Android Virtual Devices (AVDs), with the exact number determined by the batch size. Each AVD runs on an emulated device with a 1080$\times$2400 resolution, 3072 MB of memory, and 2 CPU cores. 
We train our agent for one epoch on a dataset of 436 curated GUI navigation tasks. For each task, we collect eight rollouts using 7B models and four rollouts using 32B models, with a maximum episode length of 25 steps. For a comprehensive list of all environment and training hyperparameters, please refer to the Appendix in Section~\ref{hyperparameters}.

\subsubsection{Agent Construction}
Our agent, \textbf{MobileGUI}, is built upon the \texttt{Qwen2.5-VL-7B-Instruct} and \texttt{Qwen2.5-VL-32B-Instruct} large multi-modal model. It processes both visual information from screenshots and textual task descriptions. To interact with the environment, the agent uses a structured tool-use interface, where it generates actions by calling a predefined \texttt{mobile\_use} function.

The agent is prompted to first externalize its reasoning within \texttt{<thinking>} tags and then generate a valid action call. The prompt provides the function signature, outlining the available action types and their required parameters. A detailed description of the prompt is available in Appendix~\ref{prompt_construction}. The agent's action space, as defined in the function signature, is summarized in Table~\ref{tab:action_space}.

\begin{table}[h]
\centering
\caption{Agent action space for GUI interaction.}
\label{tab:action_space}
\begin{tabular}{ll}
\toprule
\textbf{Action Type} & \textbf{Description} \\
\midrule
\texttt{click} & Tap at a specified (x, y) coordinate. \\
\texttt{swipe} & Swipe from a start coordinate to an end coordinate. \\
\texttt{type} & Input specified text into the active UI element. \\
\texttt{system\_button} & Press a system-level button (e.g., Back, Home). \\
\texttt{wait} & Pause execution for a specified number of seconds. \\
\texttt{terminate} & End the task, declaring final success or failure. \\
\texttt{answer} & Provide a textual response for question-answering tasks. \\
\bottomrule
\end{tabular}
\end{table}

\subsubsection{Baselines and Benchmarks}
We evaluate on three online GUI agent benchmarks that require agents to complete a variety of tasks within interactive environments. The evaluation spans three benchmark settings: \textbf{AndroidWorld (AW)}~\citep{rawles2024androidworld}, \textbf{Android-in-the-Wild General Tasks (AITW-Gen)}, and \textbf{Android-in-the-Wild WebShop (AITW-Web)}~\citep{zhang2024android}. Performance is measured using several metrics, including \textbf{Success Rate (SR)}, which reflects the proportion of tasks successfully completed.

Our results are compared against a range of state-of-the-art closed-source and open-source models, including GPT-4o, Claude Computer Use, and other notable open-source VLMs like Qwen2.5-VL and OS-Atlas. The detailed performance comparison is presented in the results section.

\subsection{Main Results}
We evaluate our MobileGUI-RL framework by applying it to two powerful base models, Qwen2.5-VL-7B and Qwen2.5-VL-32B, creating our MobileGUI-7B and MobileGUI-32B agents. The performance of our models against state-of-the-art closed-source and open-source baselines is detailed in Table~\ref{tab:mogui-expanded}. 

\begin{table*}[t]
\centering
\caption{Performance on GUI Agent Benchmarks. We report results across three online mobile GUI agent benchmarks, evaluating each method by Success Rate (SR).}
\vspace{-0.1in}
\begin{tabular}{lccc}
\toprule
\textbf{Models} & \textbf{AW (SR)} & \textbf{AITW-Gen (SR)} & \textbf{AITW-Web (SR)} \\
\midrule
\multicolumn{4}{l}{\textit{Closed-source Models}} \\
GPT-4o~\citep{hurst2024gpt} & 34.5 & - & - \\
Claude Computer Use~\citep{anthropic2024claude35} & 27.9 & - & - \\
\midrule
\multicolumn{4}{l}{\textit{Open-source 7B Models}} \\
OS-Genesis-7B~\citep{sun2024genesis} & - & 0.7 & 0.0 \\
OS-Atlas-7B~\citep{wu2024atlas} & - & 15.7 & 17.3 \\
Aguvis-7B~\citep{huang2024understanding} & - & 23.0 & 4.7 \\
Qwen2.5-VL-7B~\citep{bai2025qwen2} & 22.0 & 49.0 & 20.0 \\
UI-TARS-7B~\citep{qin2025ui} & 33.0 & 48.0 &16.7  \\
\textbf{MobileGUI 7B (Ours)} & 30.0 &\textbf{65.3}  & 22.7  \\
\midrule
\multicolumn{4}{l}{\textit{Open-source 32B/72B Models}} \\
Qwen2.5-VL-32B~\citep{bai2025qwen2} & 31.5 & 42.7 & 24.7 \\
Qwen2.5-VL-72B~\citep{bai2025qwen2} & \underline{35.0}  & 51.3  &\textbf{31.3} \\
Aguvis-72B~\citep{huang2024understanding} & 26.1 & -  & - \\
\textbf{MobileGUI 32B (Ours)} & \textbf{44.8} & \underline{58.0} & \underline{30.7} \\
\bottomrule
\end{tabular}
\label{tab:mogui-expanded}
\end{table*}

As presented in Table~\ref{tab:mogui-expanded}, our MobileGUI-RL framework delivers substantial performance enhancements to the base models across all three benchmarks. The results underscore the efficacy of online reinforcement learning for improving GUI navigation capabilities in large multi-modal models.

Our smaller model, \textbf{MobileGUI-7B}, demonstrates significant gains over its base model, Qwen2.5-VL-7B. The Success Rate (SR) on AndroidWorld (AW) improves from 22.0\% to 30.0\%, and most notably, we see a remarkable jump on Android-in-the-Wild General (AITW-Gen) tasks from 49.0\% to 65.3\%. This represents a 16.3 point improvement. While UI-TARS-7B shows a slightly higher SR on AW, our model's dominant performance on AITW-Gen highlights its superior ability to generalize to diverse, real-world scenarios.

The most compelling results are observed with our larger model. \textbf{MobileGUI-32B} boosts the performance of its base model, Qwen2.5-VL-32B, by 13.3 points. On the challenging AndroidWorld benchmark, our model achieves an SR of \textbf{44.8\%}, decisively outperforming all other baseline models, including the leading closed-source model GPT-4o (34.5\%) and the much larger Qwen2.5-VL-72B (35.0\%). This demonstrates that our RL fine-tuning method is not only effective but also highly efficient, enabling a 32B model to surpass a 72B model from the same family. In addition, MobileGUI-32B achieves strong performance on AITW-Gen (58.0\%) and AITW-Web (30.7\%), showing consistent and robust gains across diverse task distributions.

In summary, our MobileGUI-RL framework makes consistent and significant performance gains. The improvements are particularly pronounced in the larger model, suggesting that our online learning approach effectively refines the model's existing capabilities to master complex GUI interaction tasks. The state-of-the-art performance of MobileGUI-32B, especially on AndroidWorld, validates our method as a powerful approach for developing highly capable and efficient GUI agents.

\subsection{Ablation Study}

We conduct a series of ablation studies to systematically evaluate the contribution of each key component within our MobileGUI-RL framework. Specifically, we investigate the impact of: (1) our text-based world model for task filtering, (2) the implicit curriculum learning derived from it, and (3) our multi-component decaying reward function. We benchmark all variants on the Android World (AW) dataset, and the results are summarized in Table~\ref{tab:ablation}.

\begin{table}[h]
\centering
\caption{Ablation study of our key components on the Android World (AW) benchmark. We report the task success rate (\%). Our full MobileGUI-RL model significantly outperforms variants where a key component is removed, demonstrating the effectiveness of each design choice.}
\label{tab:ablation}
\begin{tabular}{@{}lcc@{}}
\toprule
\textbf{Configuration} & \textbf{7B Model (AW \%)} & \textbf{32B Model (AW \%)} \\ \midrule
MobileGUI-RL (Full Model) & 30.0 & 44.8 \\
\quad w/o Task Filtering & 28.5 & 41.0 \\
\quad w/o Curriculum Learning & 25.0 & 34.0 \\
\quad w/o Decaying Reward & 23.5 & 35.5 \\ \bottomrule
\end{tabular}
\end{table}

\paragraph{The Effect of Task Filtering.}
To validate the effectiveness of our task filtering mechanism, we compare our full model against a variant trained on the complete, unfiltered set of synthetically generated tasks. Our self-exploration phase initially produced 1251 candidate tasks. Our text-based world model filter pruned this set to 436 tasks deemed solvable and unambiguous. As shown in Table~\ref{tab:ablation}, removing this filter leads to a substantial performance degradation of 1.5 and 3.8 percentage points for the 7B and 32B models, respectively. This highlights the critical importance of filtering. Training on the unfiltered set exposes the agent to a high volume of low-quality or unsolvable tasks, which introduces significant noise into the learning process. This forces the agent to waste computational resources on unproductive trajectories, ultimately destabilizing policy optimization and resulting in a less capable final agent.

\paragraph{The Effect of Curriculum Learning.}
Our task generation pipeline implicitly creates a curriculum by estimating task complexity via the number of steps required for completion in the text-based world model. To ablate its effect, we trained a model on the same filtered task set but sampled tasks uniformly at random, removing the complexity-based ordering. The results, summarized in Table~\ref{tab:ablation}, demonstrate a substantial performance drop when the curriculum is removed. The 7B model's success rate falls by 5 points (from 30.0\% to 25.0\%), and the 32B model's performance drops by a significant 10.8 points (from 44.8\% to 34.0\%). This highlights the curriculum's critical role in achieving high final performance.
\begin{figure}[h!]
    \centering
    \begin{subfigure}[b]{0.48\textwidth}
        \centering
        \includegraphics[width=\textwidth]{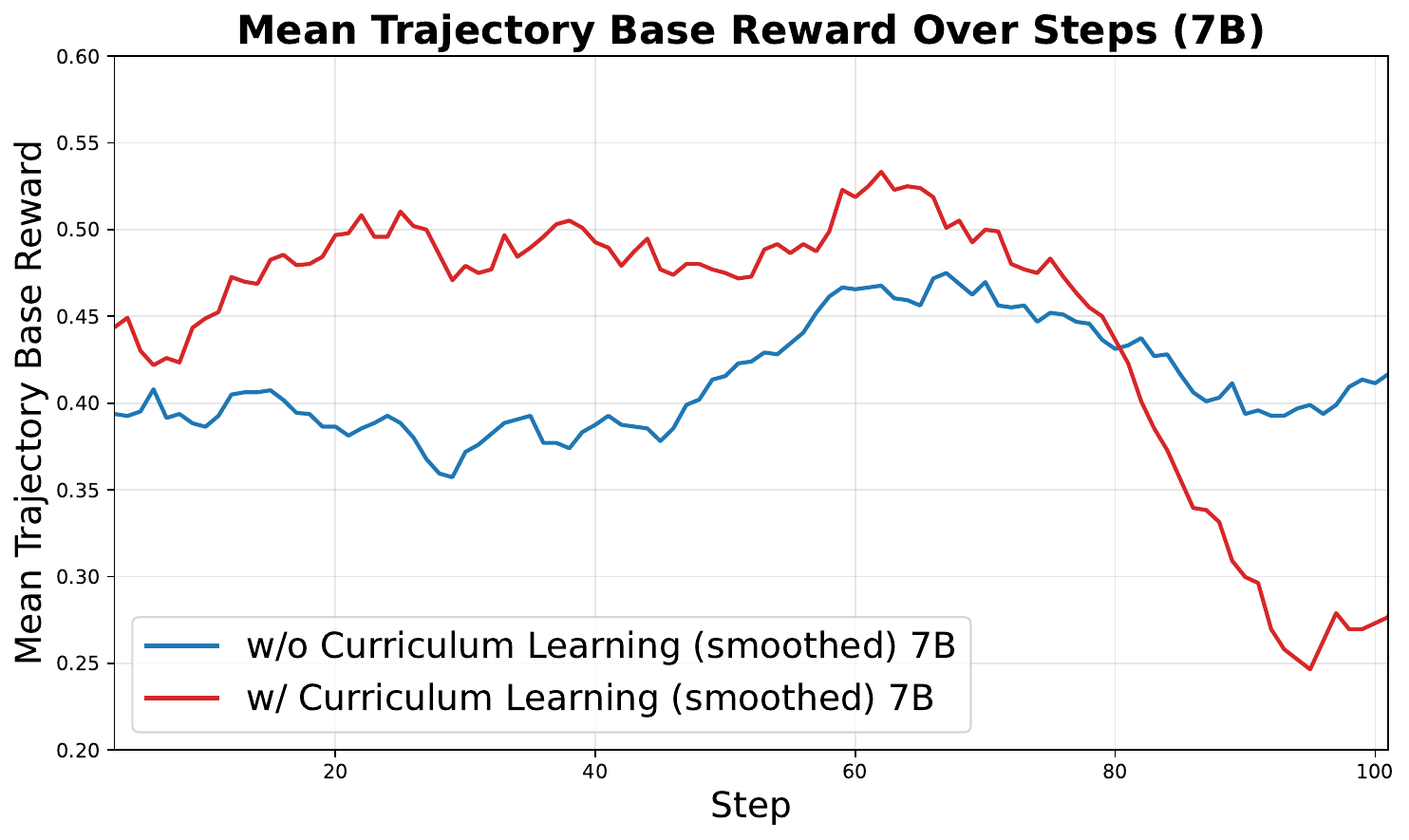}
        \vspace{-0.2in}    
        \caption{Mean Reward (7B)}
        \label{fig:reward_7b}
    \end{subfigure}
    \hfill
    \begin{subfigure}[b]{0.48\textwidth}
        \centering
        \includegraphics[width=\textwidth]{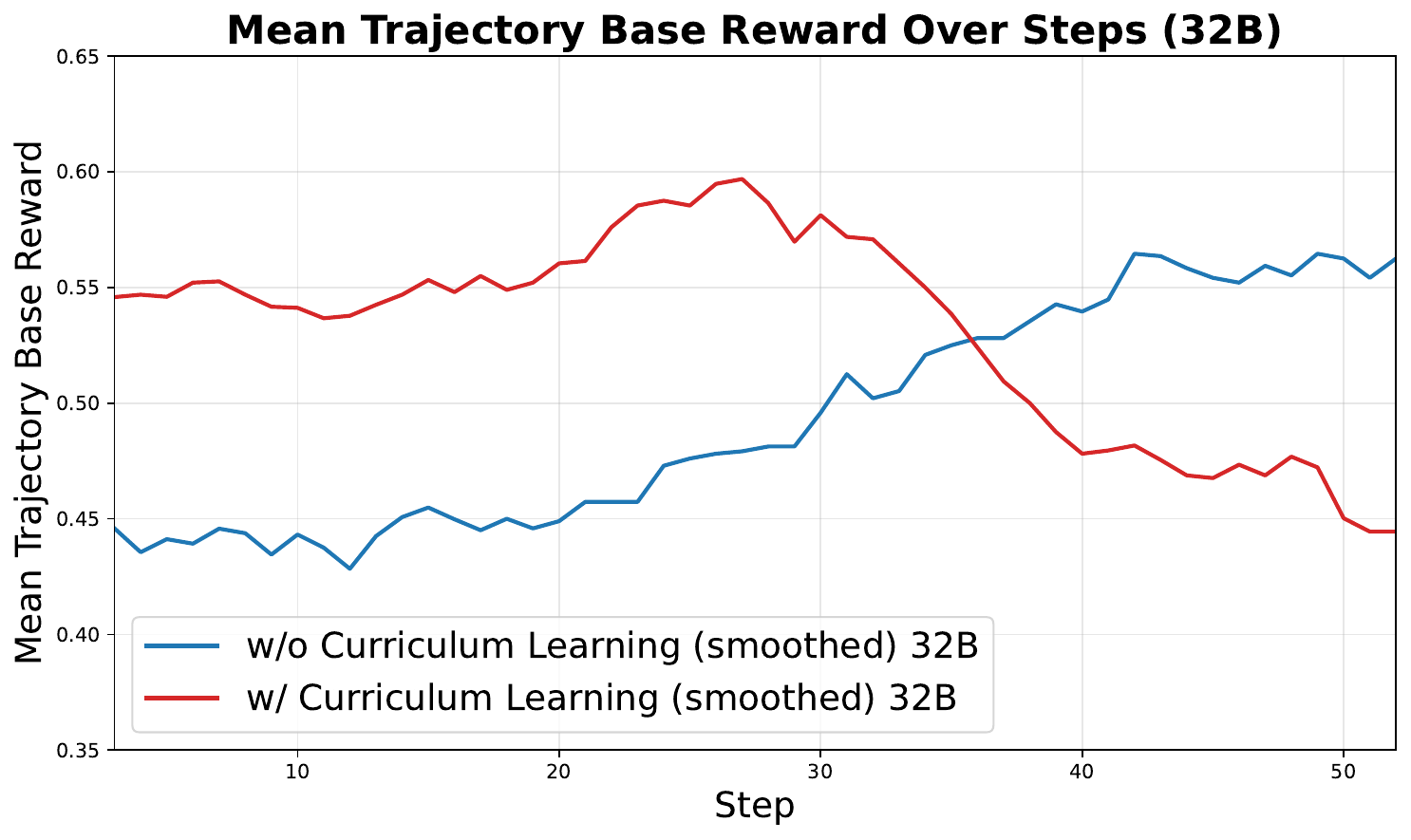}
        \vspace{-0.2in}    
        \caption{Mean Reward (32B)}
        \label{fig:reward_32b}
    \end{subfigure}
     
    \vspace{0.2cm} 
    
    \begin{subfigure}[b]{0.48\textwidth}
        \centering
        \includegraphics[width=\textwidth]{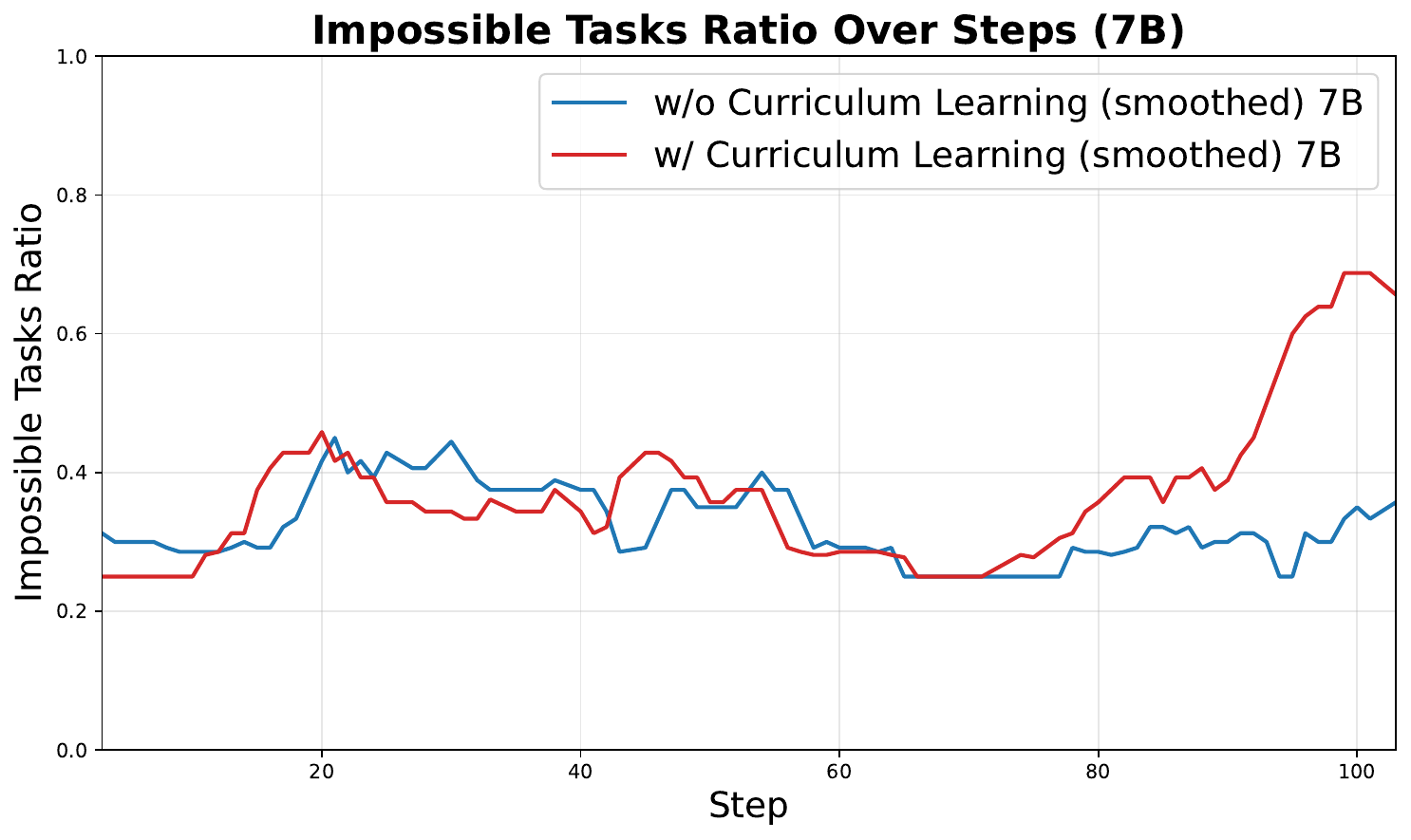}
        \vspace{-0.2in}    
        \caption{Impossible Tasks Ratio (7B)}
        \label{fig:impossible_7b}
    \end{subfigure}
    \hfill
    \begin{subfigure}[b]{0.48\textwidth}
        \centering
        \includegraphics[width=\textwidth]{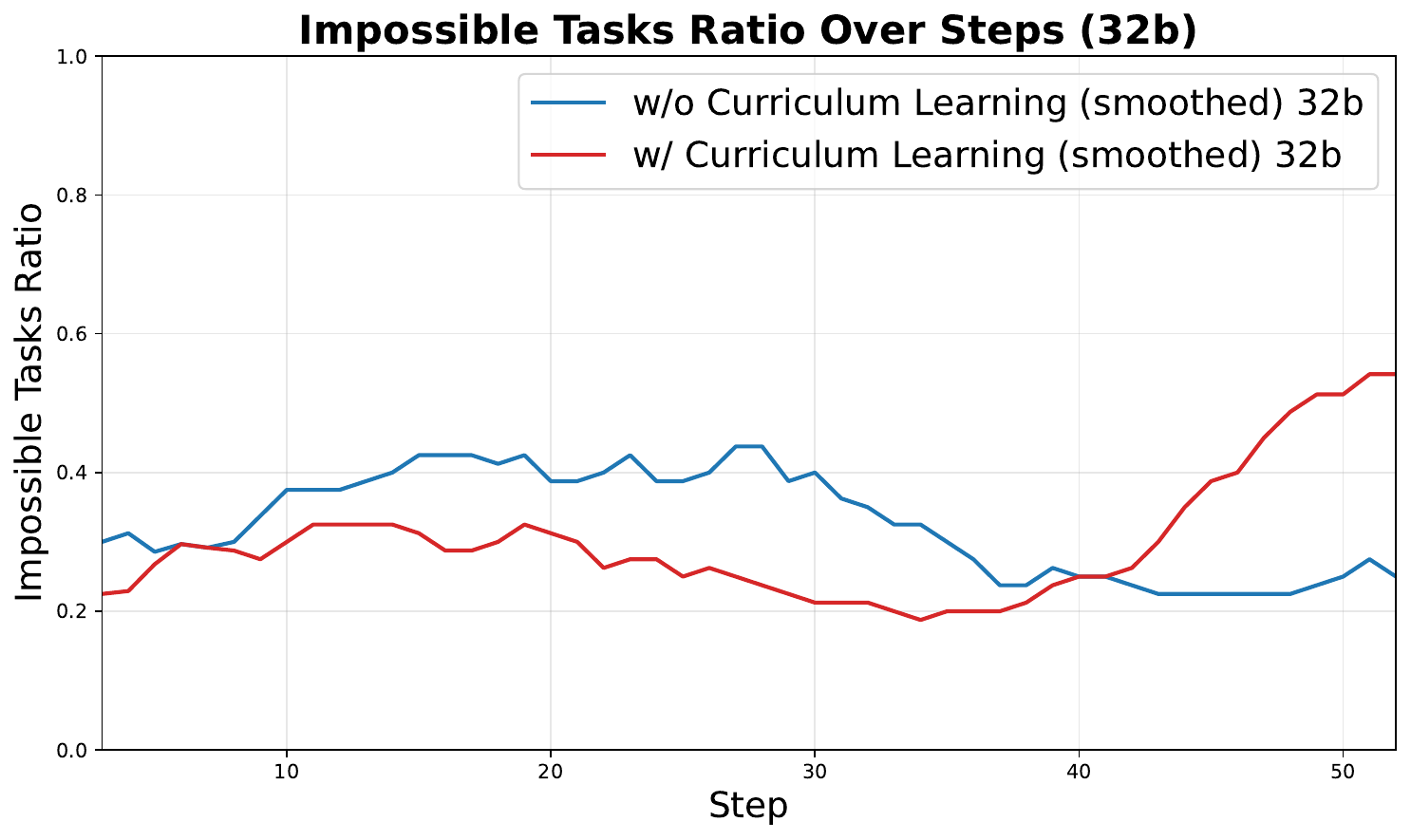}
        \vspace{-0.2in}    
        \caption{Impossible Tasks Ratio (32B)}
        \label{fig:impossible_32b}
    \end{subfigure}
    \vspace{-0.1in}    
    \caption{Training dynamics with and without curriculum learning for 7B and 32B models. The top row shows the mean trajectory base reward, and the bottom row shows the ratio of impossible tasks encountered. With curriculum learning (red), the reward first rises on easy tasks and then falls as the curriculum introduces harder tasks, which is corroborated by the rising impossible task ratio. This structured approach leads to better final performance than training without a curriculum (blue).}
    \label{fig:curriculum_plots}
\end{figure}

The training dynamics, illustrated in Figure~\ref{fig:curriculum_plots}, provide deeper insight into why the curriculum is so effective. By starting with simpler tasks, the curriculum-based approach allows the agent to first build a robust foundation of basic interaction skills. It then progressively introduces more complex tasks that challenge the agent to develop sophisticated, multi-step reasoning.

This structured learning process is evident in the training curves. For models trained with curriculum learning (red lines), the mean reward (Figures~\ref{fig:reward_7b} and \ref{fig:reward_32b}) initially rises as the agent masters the beginning set of easier tasks. Subsequently, the reward curve trends downward. This decline does not indicate that the model is forgetting or degrading. Instead, it reflects the nature of the curriculum, which introduces progressively harder tasks in the later stages of training. As shown in Figures~\ref{fig:impossible_7b} and \ref{fig:impossible_32b}, the ratio of impossible tasks, those that remain unsolved across all attempts, increases towards the end of training, naturally leading to a lower average success rate and reward on these more difficult task distributions.

In contrast, the models trained without a curriculum (blue lines) are exposed to a mix of easy and hard tasks from the start, resulting in a more stationary reward signal throughout training. Although the training curves for the non-curriculum models might appear more stable, their lower final performance on the benchmark (Table~\ref{tab:ablation}) confirms that this uniform sampling is less effective. Our curriculum-based method proves to be more sample-efficient and ultimately leads to a more capable agent with higher final performance.

\paragraph{The Effect of Decaying Reward.}
Finally, we evaluate the effectiveness of our multi-component reward design, focusing on the exponential decay factor that encourages efficiency. We compare our full model against a variant using a simple binary reward ($r=1$ for success, $r=0$ otherwise). Removing the decaying component leads to significant performance drops of 6.5 and 9.3 points for the 7B and 32B models, respectively, highlighting the limitations of sparse, binary rewards in complex GUI navigation tasks.
The decaying reward plays two key roles. First, it introduces reward variance among successful trajectories, motivating the agent to seek not just correct, but efficient solutions. Second, it mitigates a common failure mode in GRPO-style algorithms: when all trajectories in a batch succeed with identical rewards, the normalized advantage becomes zero, halting learning. By preserving gradient signals and rewarding efficiency, our reward formulation provides dense, informative feedback that is critical for effective policy optimization.



\section{Conclusion}

In this work, we introduced MobileGUI-RL, a reinforcement learning framework designed to train GUI agents in online environments. By creating an interactive environment, our framework enables agents to adapt to the dynamic and unpredictable nature of mobile UI interfaces. We addressed the challenge of data generation through a synthetic task pipeline that combines self-exploration with a text-based world model for curriculum filtering. Furthermore, our MobGRPO algorithm incorporates a trajectory-aware advantage and a multi-component reward, effectively optimizing for both task success and interaction efficiency.

Our experiments demonstrate the effectiveness of this approach. By fine-tuning strong base models, our MobileGUI agents achieved significant performance gains across several challenging GUI benchmarks. Notably, MobileGUI-32B surpasses both its base model and leading closed-source competitors. These results provide strong evidence that online reinforcement learning from trajectory-level feedback is a powerful paradigm for developing more robust and capable GUI agents.


\section{Future Directions}
Building upon this work, future research will focus on several key areas to advance mobile GUI agents. A primary direction is enhancing task complexity beyond self-exploration by generating more realistic, long-horizon tasks through methods like human-in-the-loop curation and hierarchical decomposition. To train on these more challenging tasks effectively, we aim to refine our reward design by shifting from sparse, trajectory-level feedback to more granular, step-wise supervision. Trajectory-level rewards can introduce ambiguous learning signals, particularly when both successful and failed trajectories share common steps, making it difficult for the agent to discern which actions contributed to success. Another leap forward will involve developing visual world models that enable agents to perform multi-step lookahead planning by predicting future screens, drastically improving error correction and strategic execution. Finally, these advancements will pave the way for true personalization through on-device continual adaptation, allowing agents to learn from a specific user's patterns and preferences for a more integrated and effective experience.




\bibliography{ref, agent_ref}

\begin{thebibliography}{36}
\providecommand{\natexlab}[1]{#1}
\providecommand{\url}[1]{\texttt{#1}}
\expandafter\ifx\csname urlstyle\endcsname\relax
  \providecommand{\doi}[1]{doi: #1}\else
  \providecommand{\doi}{doi: \begingroup \urlstyle{rm}\Url}\fi

\bibitem[{Android Developers}(2024)]{android_emulator}
{Android Developers}.
\newblock {Run apps on the Android Emulator}.
\newblock \url{https://developer.android.com/studio/run/emulator}, 2024.
\newblock Accessed: 2025-06-23.

\bibitem[{Anthropic}(2024)]{anthropic2024claude35}
{Anthropic}.
\newblock Introducing computer use, a new claude 3.5 sonnet, and claude 3.5 haiku.
\newblock \url{https://www.anthropic.com/news/3-5-models-and-computer-use}, 2024.
\newblock Accessed: 2025-06-23.

\bibitem[Anthropic(2025)]{anthropic2025claude}
Anthropic.
\newblock Introducing claude 3.5 sonnet.
\newblock \emph{https://www.anthropic.com/news/claude-3-5-sonnet}, 2025.

\bibitem[Bai et~al.(2024)Bai, Zhou, Pan, Cemri, Suhr, Levine, and Kumar]{bai2024digirl}
Hao Bai, Yifei Zhou, Jiayi Pan, Mert Cemri, Alane Suhr, Sergey Levine, and Aviral Kumar.
\newblock Digirl: Training in-the-wild device-control agents with autonomous reinforcement learning.
\newblock \emph{Advances in Neural Information Processing Systems}, 37:\penalty0 12461--12495, 2024.

\bibitem[Bai et~al.(2025)Bai, Chen, Liu, Wang, Ge, Song, Dang, Wang, Wang, Tang, et~al.]{bai2025qwen2}
Shuai Bai, Keqin Chen, Xuejing Liu, Jialin Wang, Wenbin Ge, Sibo Song, Kai Dang, Peng Wang, Shijie Wang, Jun Tang, et~al.
\newblock Qwen2. 5-vl technical report.
\newblock \emph{arXiv preprint arXiv:2502.13923}, 2025.

\bibitem[Fang et~al.(2025)Fang, Zhang, Zhang, Ma, Yu, Mi, and Yu]{fang2025webevolver}
Tianqing Fang, Hongming Zhang, Zhisong Zhang, Kaixin Ma, Wenhao Yu, Haitao Mi, and Dong Yu.
\newblock Webevolver: Enhancing web agent self-improvement with coevolving world model.
\newblock \emph{arXiv preprint arXiv:2504.21024}, 2025.

\bibitem[Feng et~al.(2025)Feng, Xue, Liu, and An]{feng2025group}
Lang Feng, Zhenghai Xue, Tingcong Liu, and Bo~An.
\newblock Group-in-group policy optimization for llm agent training.
\newblock \emph{arXiv preprint arXiv:2505.10978}, 2025.

\bibitem[Gou et~al.(2025)Gou, Wang, Zheng, Xie, Chang, Shu, Sun, and Su]{gou2025navigating}
Boyu Gou, Ruohan Wang, Boyuan Zheng, Yanan Xie, Cheng Chang, Yiheng Shu, Huan Sun, and Yu~Su.
\newblock Navigating the digital world as humans do: Universal visual grounding for gui agents.
\newblock \emph{International Conference for Learning Representation (ICLR)}, 2025.

\bibitem[Guo et~al.(2025)Guo, Yang, Zhang, Song, Zhang, Xu, Zhu, Ma, Wang, Bi, et~al.]{guo2025deepseek}
Daya Guo, Dejian Yang, Haowei Zhang, Junxiao Song, Ruoyu Zhang, Runxin Xu, Qihao Zhu, Shirong Ma, Peiyi Wang, Xiao Bi, et~al.
\newblock Deepseek-r1: Incentivizing reasoning capability in llms via reinforcement learning.
\newblock \emph{arXiv preprint arXiv:2501.12948}, 2025.

\bibitem[Gur et~al.(2023)Gur, Furuta, Huang, Safdari, Matsuo, Eck, and Faust]{gur2023real}
Izzeddin Gur, Hiroki Furuta, Austin Huang, Mustafa Safdari, Yutaka Matsuo, Douglas Eck, and Aleksandra Faust.
\newblock A real-world webagent with planning, long context understanding, and program synthesis.
\newblock \emph{arXiv preprint arXiv:2307.12856}, 2023.

\bibitem[Hong et~al.(2024)Hong, Wang, Lv, Xu, Yu, Ji, Wang, Wang, Dong, Ding, et~al.]{hong2024cogagent}
Wenyi Hong, Weihan Wang, Qingsong Lv, Jiazheng Xu, Wenmeng Yu, Junhui Ji, Yan Wang, Zihan Wang, Yuxiao Dong, Ming Ding, et~al.
\newblock Cogagent: A visual language model for gui agents.
\newblock In \emph{Proceedings of the IEEE/CVF Conference on Computer Vision and Pattern Recognition}, pp.\  14281--14290, 2024.

\bibitem[Hu et~al.(2025)Hu, Fang, Zhang, Ma, Zhang, Zhou, Zhang, Mi, Yu, and King]{hu2025webcot}
Minda Hu, Tianqing Fang, Jianshu Zhang, Junyu Ma, Zhisong Zhang, Jingyan Zhou, Hongming Zhang, Haitao Mi, Dong Yu, and Irwin King.
\newblock Webcot: Enhancing web agent reasoning by reconstructing chain-of-thought in reflection, branching, and rollback.
\newblock \emph{arXiv preprint arXiv:2505.20013}, 2025.

\bibitem[Huang et~al.(2024)Huang, Liu, Chen, Wang, Wang, Lian, Wang, Tang, and Chen]{huang2024understanding}
Xu~Huang, Weiwen Liu, Xiaolong Chen, Xingmei Wang, Hao Wang, Defu Lian, Yasheng Wang, Ruiming Tang, and Enhong Chen.
\newblock Understanding the planning of llm agents: A survey.
\newblock \emph{arXiv preprint arXiv:2402.02716}, 2024.

\bibitem[Hurst et~al.(2024)Hurst, Lerer, Goucher, Perelman, Ramesh, Clark, Ostrow, Welihinda, Hayes, Radford, et~al.]{hurst2024gpt}
Aaron Hurst, Adam Lerer, Adam~P Goucher, Adam Perelman, Aditya Ramesh, Aidan Clark, AJ~Ostrow, Akila Welihinda, Alan Hayes, Alec Radford, et~al.
\newblock Gpt-4o system card.
\newblock \emph{arXiv preprint arXiv:2410.21276}, 2024.

\bibitem[Liu et~al.(2025)Liu, Li, Xie, Hu, Han, Zhang, Yang, and Wu]{liu2025infigui}
Yuhang Liu, Pengxiang Li, Congkai Xie, Xavier Hu, Xiaotian Han, Shengyu Zhang, Hongxia Yang, and Fei Wu.
\newblock Infigui-r1: Advancing multimodal gui agents from reactive actors to deliberative reasoners.
\newblock \emph{arXiv preprint arXiv:2504.14239}, 2025.

\bibitem[Lu et~al.(2025{\natexlab{a}})Lu, Zhong, Liu, Fu, and Jia]{lu2025arpo}
Fanbin Lu, Zhisheng Zhong, Shu Liu, Chi-Wing Fu, and Jiaya Jia.
\newblock Arpo: End-to-end policy optimization for gui agents with experience replay.
\newblock \emph{arXiv preprint arXiv:2505.16282}, 2025{\natexlab{a}}.

\bibitem[Lu et~al.(2025{\natexlab{b}})Lu, Chai, Guo, Yin, Liu, Wang, Xiao, Ren, Xiong, and Li]{lu2025ui}
Zhengxi Lu, Yuxiang Chai, Yaxuan Guo, Xi~Yin, Liang Liu, Hao Wang, Han Xiao, Shuai Ren, Guanjing Xiong, and Hongsheng Li.
\newblock Ui-r1: Enhancing action prediction of gui agents by reinforcement learning.
\newblock \emph{arXiv preprint arXiv:2503.21620}, 2025{\natexlab{b}}.

\bibitem[Luo et~al.(2025)Luo, Wang, He, and Xia]{luo2025gui}
Run Luo, Lu~Wang, Wanwei He, and Xiaobo Xia.
\newblock Gui-r1: A generalist r1-style vision-language action model for gui agents.
\newblock \emph{arXiv preprint arXiv:2504.10458}, 2025.

\bibitem[Qin et~al.(2025)Qin, Ye, Fang, Wang, Liang, Tian, Zhang, Li, Li, Huang, et~al.]{qin2025ui}
Yujia Qin, Yining Ye, Junjie Fang, Haoming Wang, Shihao Liang, Shizuo Tian, Junda Zhang, Jiahao Li, Yunxin Li, Shijue Huang, et~al.
\newblock Ui-tars: Pioneering automated gui interaction with native agents.
\newblock \emph{arXiv preprint arXiv:2501.12326}, 2025.

\bibitem[Rawles et~al.(2024)Rawles, Clinckemaillie, Chang, Waltz, Lau, Fair, Li, Bishop, Li, Campbell-Ajala, et~al.]{rawles2024androidworld}
Christopher Rawles, Sarah Clinckemaillie, Yifan Chang, Jonathan Waltz, Gabrielle Lau, Marybeth Fair, Alice Li, William Bishop, Wei Li, Folawiyo Campbell-Ajala, et~al.
\newblock Androidworld: A dynamic benchmarking environment for autonomous agents.
\newblock \emph{arXiv preprint arXiv:2405.14573}, 2024.

\bibitem[Schulman et~al.(2017)Schulman, Wolski, Dhariwal, Radford, and Klimov]{schulman2017proximal}
John Schulman, Filip Wolski, Prafulla Dhariwal, Alec Radford, and Oleg Klimov.
\newblock Proximal policy optimization algorithms.
\newblock \emph{arXiv preprint arXiv:1707.06347}, 2017.

\bibitem[Shao et~al.(2024)Shao, Wang, Zhu, Xu, Song, Bi, Zhang, Zhang, Li, Wu, et~al.]{shao2024deepseekmath}
Zhihong Shao, Peiyi Wang, Qihao Zhu, Runxin Xu, Junxiao Song, Xiao Bi, Haowei Zhang, Mingchuan Zhang, YK~Li, Y~Wu, et~al.
\newblock Deepseekmath: Pushing the limits of mathematical reasoning in open language models.
\newblock \emph{arXiv preprint arXiv:2402.03300}, 2024.

\bibitem[Sun et~al.(2024)Sun, Cheng, Ding, Jin, Wang, Xu, Wu, Jia, Chen, Liu, et~al.]{sun2024genesis}
Qiushi Sun, Kanzhi Cheng, Zichen Ding, Chuanyang Jin, Yian Wang, Fangzhi Xu, Zhenyu Wu, Chengyou Jia, Liheng Chen, Zhoumianze Liu, et~al.
\newblock Os-genesis: Automating gui agent trajectory construction via reverse task synthesis.
\newblock \emph{arXiv preprint arXiv:2412.19723}, 2024.

\bibitem[Sun et~al.(2025)Sun, Cheng, Ding, Jin, Wang, Xu, Wu, Jia, Chen, Liu, et~al.]{sun2025genesis}
Qiushi Sun, Kanzhi Cheng, Zichen Ding, Chuanyang Jin, Yian Wang, Fangzhi Xu, Zhenyu Wu, Chengyou Jia, Liheng Chen, Zhoumianze Liu, et~al.
\newblock Os-genesis: Automating gui agent trajectory construction via reverse task synthesis.
\newblock In \emph{The 63rd Annual Meeting of the Association for Computational Linguistics}, 2025.

\bibitem[Wang et~al.(2024)Wang, Liu, Chen, Zhou, Gan, Zeng, Che, Yu, Hao, Shao, et~al.]{wang2024gui}
Shuai Wang, Weiwen Liu, Jingxuan Chen, Yuqi Zhou, Weinan Gan, Xingshan Zeng, Yuhan Che, Shuai Yu, Xinlong Hao, Kun Shao, et~al.
\newblock Gui agents with foundation models: A comprehensive survey.
\newblock \emph{arXiv preprint arXiv:2411.04890}, 2024.

\bibitem[Wang et~al.(2025)Wang, Wu, Liu, HAO, Wang, and Shao]{wang2025distrl}
Taiyi Wang, Zhihao Wu, Jianheng Liu, Jianye HAO, Jun Wang, and Kun Shao.
\newblock Dist{RL}: An asynchronous distributed reinforcement learning framework for on-device control agent.
\newblock In \emph{The Thirteenth International Conference on Learning Representations}, 2025.
\newblock URL \url{https://openreview.net/forum?id=LPG8pPSfQD}.

\bibitem[Wei et~al.(2025)Wei, Yao, Liu, Zhang, Lu, Qiu, Yu, Xu, Zhang, Yin, et~al.]{wei2025webagent}
Zhepei Wei, Wenlin Yao, Yao Liu, Weizhi Zhang, Qin Lu, Liang Qiu, Changlong Yu, Puyang Xu, Chao Zhang, Bing Yin, et~al.
\newblock Webagent-r1: Training web agents via end-to-end multi-turn reinforcement learning.
\newblock \emph{arXiv preprint arXiv:2505.16421}, 2025.

\bibitem[Wu et~al.(2024)Wu, Wu, Xu, Wang, Sun, Jia, Cheng, Ding, Chen, Liang, et~al.]{wu2024atlas}
Zhiyong Wu, Zhenyu Wu, Fangzhi Xu, Yian Wang, Qiushi Sun, Chengyou Jia, Kanzhi Cheng, Zichen Ding, Liheng Chen, Paul~Pu Liang, et~al.
\newblock Os-atlas: A foundation action model for generalist gui agents.
\newblock \emph{arXiv preprint arXiv:2410.23218}, 2024.

\bibitem[Xu et~al.(2025)Xu, Wang, Wang, Lu, Xie, Saha, Sahoo, Yu, and Xiong]{xu2025aguvis}
Yiheng Xu, Zekun Wang, Junli Wang, Dunjie Lu, Tianbao Xie, Amrita Saha, Doyen Sahoo, Tao Yu, and Caiming Xiong.
\newblock Aguvis: Unified pure vision agents for autonomous gui interaction.
\newblock 2025.

\bibitem[Yang et~al.(2025)Yang, Su, Liu, Dong, Yu, Su, Wang, Liu, Zhu, Li, et~al.]{yang2025zerogui}
Chenyu Yang, Shiqian Su, Shi Liu, Xuan Dong, Yue Yu, Weijie Su, Xuehui Wang, Zhaoyang Liu, Jinguo Zhu, Hao Li, et~al.
\newblock Zerogui: Automating online gui learning at zero human cost.
\newblock \emph{arXiv preprint arXiv:2505.23762}, 2025.

\bibitem[Yu et~al.(2025)Yu, Zhang, Zhu, Yuan, Zuo, Yue, Dai, Fan, Liu, Liu, et~al.]{yu2025dapo}
Qiying Yu, Zheng Zhang, Ruofei Zhu, Yufeng Yuan, Xiaochen Zuo, Yu~Yue, Weinan Dai, Tiantian Fan, Gaohong Liu, Lingjun Liu, et~al.
\newblock Dapo: An open-source llm reinforcement learning system at scale.
\newblock \emph{arXiv preprint arXiv:2503.14476}, 2025.

\bibitem[Zhang et~al.(2024{\natexlab{a}})Zhang, He, Qian, Li, Li, Qin, Kang, Ma, Liu, Lin, et~al.]{zhang2024large}
Chaoyun Zhang, Shilin He, Jiaxu Qian, Bowen Li, Liqun Li, Si~Qin, Yu~Kang, Minghua Ma, Guyue Liu, Qingwei Lin, et~al.
\newblock Large language model-brained gui agents: A survey.
\newblock \emph{arXiv preprint arXiv:2411.18279}, 2024{\natexlab{a}}.

\bibitem[Zhang et~al.(2024{\natexlab{b}})Zhang, Wu, Teng, Liao, Xu, Xiao, Wei, and Tang]{zhang2024android}
Jiwen Zhang, Jihao Wu, Yihua Teng, Minghui Liao, Nuo Xu, Xiao Xiao, Zhongyu Wei, and Duyu Tang.
\newblock Android in the zoo: Chain-of-action-thought for gui agents.
\newblock \emph{arXiv preprint arXiv:2403.02713}, 2024{\natexlab{b}}.

\bibitem[Zhang et~al.(2025)Zhang, Fang, Ma, Yu, Zhang, Mi, and Yu]{zhang2025enhancing}
Zhisong Zhang, Tianqing Fang, Kaixin Ma, Wenhao Yu, Hongming Zhang, Haitao Mi, and Dong Yu.
\newblock Enhancing web agents with explicit rollback mechanisms.
\newblock \emph{arXiv preprint arXiv:2504.11788}, 2025.

\bibitem[Zheng et~al.(2024)Zheng, Gou, Kil, Sun, and Su]{zheng2024gpt}
Boyuan Zheng, Boyu Gou, Jihyung Kil, Huan Sun, and Yu~Su.
\newblock Gpt-4v (ision) is a generalist web agent, if grounded.
\newblock \emph{arXiv preprint arXiv:2401.01614}, 2024.

\bibitem[Zhou et~al.(2025)Zhou, Dai, Wang, Zhou, Jia, et~al.]{zhou2025gui}
Yuqi Zhou, Sunhao Dai, Shuai Wang, Kaiwen Zhou, Qinqlin Jia, et~al.
\newblock Gui-g1: Understanding r1-zero-like training for visual grounding in gui agents.
\newblock \emph{arXiv preprint arXiv:2505.15810}, 2025.

\end{thebibliography}
\bibliographystyle{colm2024_conference}
\newpage
\appendix

\section{Detailed Training Configuration}
\label{hyperparameters}
\subsection{Environment Setup}

\subsubsection{Android Emulator Configuration}
\begin{itemize}
    \item \textbf{Base AVD Name}: AndroidWorldAvd
    \item \textbf{Emulator Instances}: Dynamically scaled based on batch size
    \item \textbf{Screen Resolution}: 1080×2400 pixels
    \item \textbf{Memory Allocation}: 3072 MB per emulator
    \item \textbf{CPU Cores}: 2 cores per emulator
    \item \textbf{GPU Acceleration}: Auto mode
\end{itemize}

\subsection{Model and Training Hyperparameters}

\subsubsection{Model Configuration}
\begin{itemize}
    \item \textbf{Base Model}: Qwen2.5-VL-7B-Instruct and Qwen2.5-VL-32B-Instruct
    \item \textbf{Attention Implementation}: Flash Attention 2
    \item \textbf{Gradient Checkpointing}: Enabled
    \item \textbf{Mixed Precision}: BFloat16 for parameters, FP32 for reduction
\end{itemize}

\subsubsection{GRPO Training Parameters}
\begin{itemize}
    \item \textbf{Global Batch Size}: 128
    \item \textbf{Micro Batch Size (Update)}: 4 per device
    \item \textbf{Micro Batch Size (Experience)}: 16 per device
    \item \textbf{Learning Rate}: $1 \times 10^{-6}$
    \item \textbf{Adam Betas}: (0.9, 0.999)
    \item \textbf{Weight Decay}: 0.01
    \item \textbf{Gradient Clipping}: 1.0
    \item \textbf{PPO Clip Ratio}: 0.2
    \item \textbf{Entropy Coefficient}: $1 \times 10^{-3}$
    \item \textbf{KL Penalty Coefficient}: $1 \times 10^{-2}$ (for GRPO)
    \item \textbf{PPO Epochs}: 1
    \item \textbf{Advantage Estimator}: GRPO with trajectory-based normalization
\end{itemize}

\subsubsection{Rollout Configuration}
\begin{itemize}
    \item \textbf{Temperature}: 1.0
    \item \textbf{Top-p}: 1.0
    \item \textbf{Max Response Length}: 2048 tokens
    \item \textbf{Number of Rollouts per Prompt}: 8
    \item \textbf{Maximum Steps per Episode}: 15
    \item \textbf{Tensor Parallel Size}: 2
    \item \textbf{GPU Memory Utilization}: 0.5
\end{itemize}

\section{Prompt Construction}
\label{prompt_construction}

The agent operates through a structured tool-use interface. The system prompt provides the agent with a function signature for mobile device interaction:

\begin{lstlisting}[language=json,basicstyle=\small]
{
  "type": "function",
  "function": {
    "name": "mobile_use",
    "description": "Use a touchscreen to interact...",
    "parameters": {
      "properties": {
        "action": {
          "enum": ["click", "swipe", "type", 
                   "system_button", "wait", 
                   "terminate", "answer"]
        },
        "coordinate": {"type": "array"},
        "coordinate2": {"type": "array"},
        "text": {"type": "string"},
        "time": {"type": "number"},
        "button": {"enum": ["Back", "Home", 
                           "Menu", "Enter"]},
        "status": {"enum": ["success", "failure"]}
      }
    }
  }
}
\end{lstlisting}

The agent is instructed to provide reasoning within \texttt{<thinking>} tags before each action and summarize actions within \texttt{<conclusion>} tags. Task progress is tracked by maintaining a history of previous actions and their outcomes.

The evaluator provides binary success/failure judgments along with detailed reasoning about whether all task requirements have been satisfied.

\section{Evaluation Details}
\label{sec:eval_details}

This section provides additional details on the evaluation procedures for each benchmark used in our experiments.

\paragraph{AndroidWorld} For the AndroidWorld benchmark, we utilized the official evaluation code and procedures released by the original authors~\citep{rawles2024androidworld}. This ensures that our results are directly comparable to previously reported scores on this benchmark.

\paragraph{Android-in-the-Wild (AITW)} For the AITW-Gen and AITW-Web benchmarks, we adapted the evaluation scripts originally provided in the DigiRL study~\citep{bai2024digirl}. We made several modifications to curate the datasets for our specific testing environment.
\begin{itemize}
    \item \textbf{AITW-Gen:} We manually reviewed the tasks and removed those that could not be reliably executed on our emulated Android environment. These tasks primarily involved actions such as installing specific third-party applications, which were not feasible in our sandboxed virtual devices. After this filtering process, the final AITW-Gen dataset used for our evaluation consisted of 300 unique tasks.
    \item \textbf{AITW-Web:} During our review of the WebShop tasks, we identified a significant number of duplicate entries. To create a more robust and less redundant benchmark, we performed a deduplication process, merging these similar tasks. This resulted in a final, curated AITW-Web benchmark of 150 unique tasks.
\end{itemize}
These curation steps were taken to ensure a fair and consistent evaluation of the agent's capabilities on tasks that are executable within our standardized environment.

\section{Cases}

\begin{figure}[h!]
    \centering
    \begin{subfigure}[b]{0.96\textwidth}
        \centering
        \includegraphics[width=\textwidth]{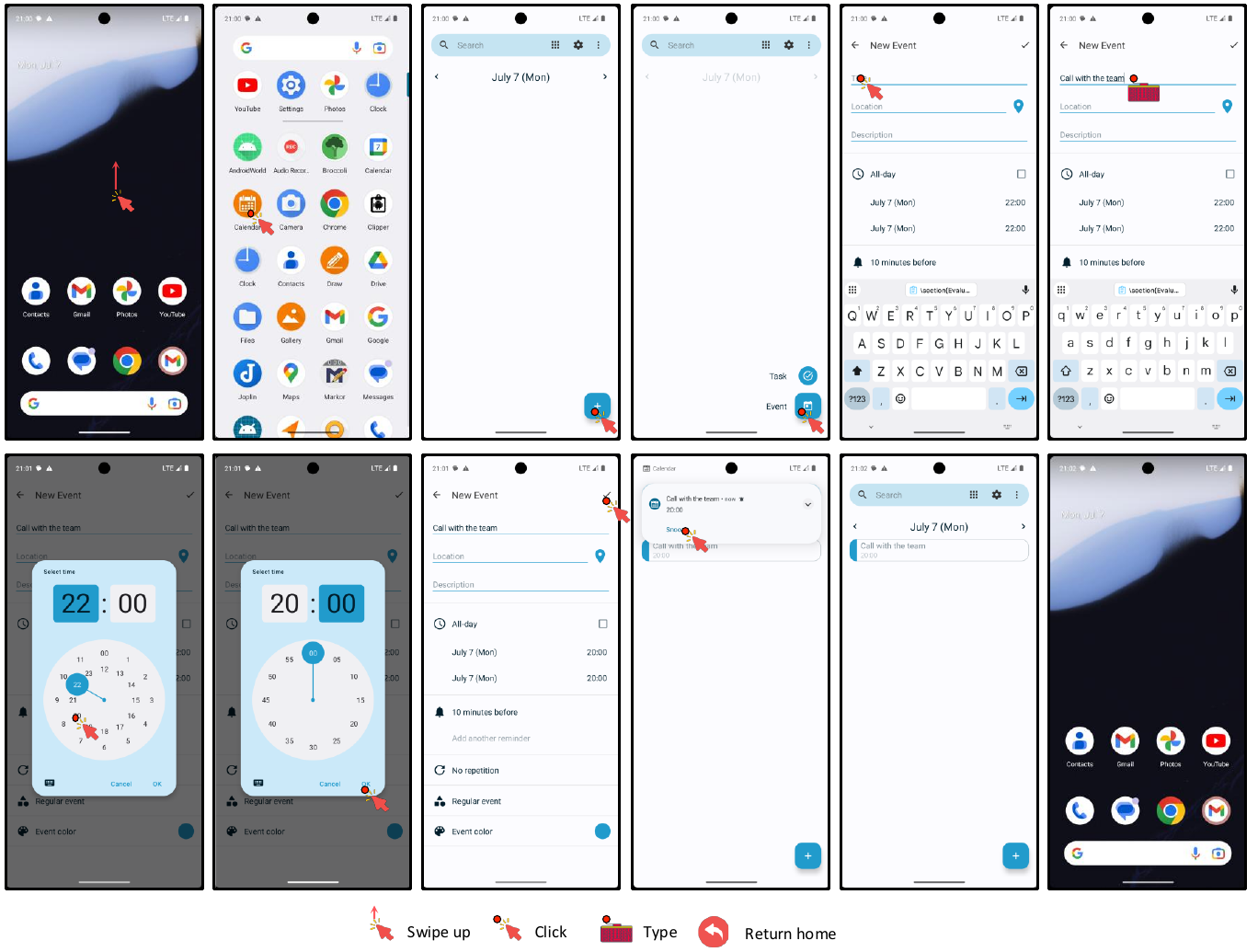}
        \label{fig:case_base}
    \end{subfigure}
\end{figure}

\begin{figure}[h!]
    \centering
    \begin{subfigure}[b]{0.96\textwidth}
        \centering
        \includegraphics[width=\textwidth]{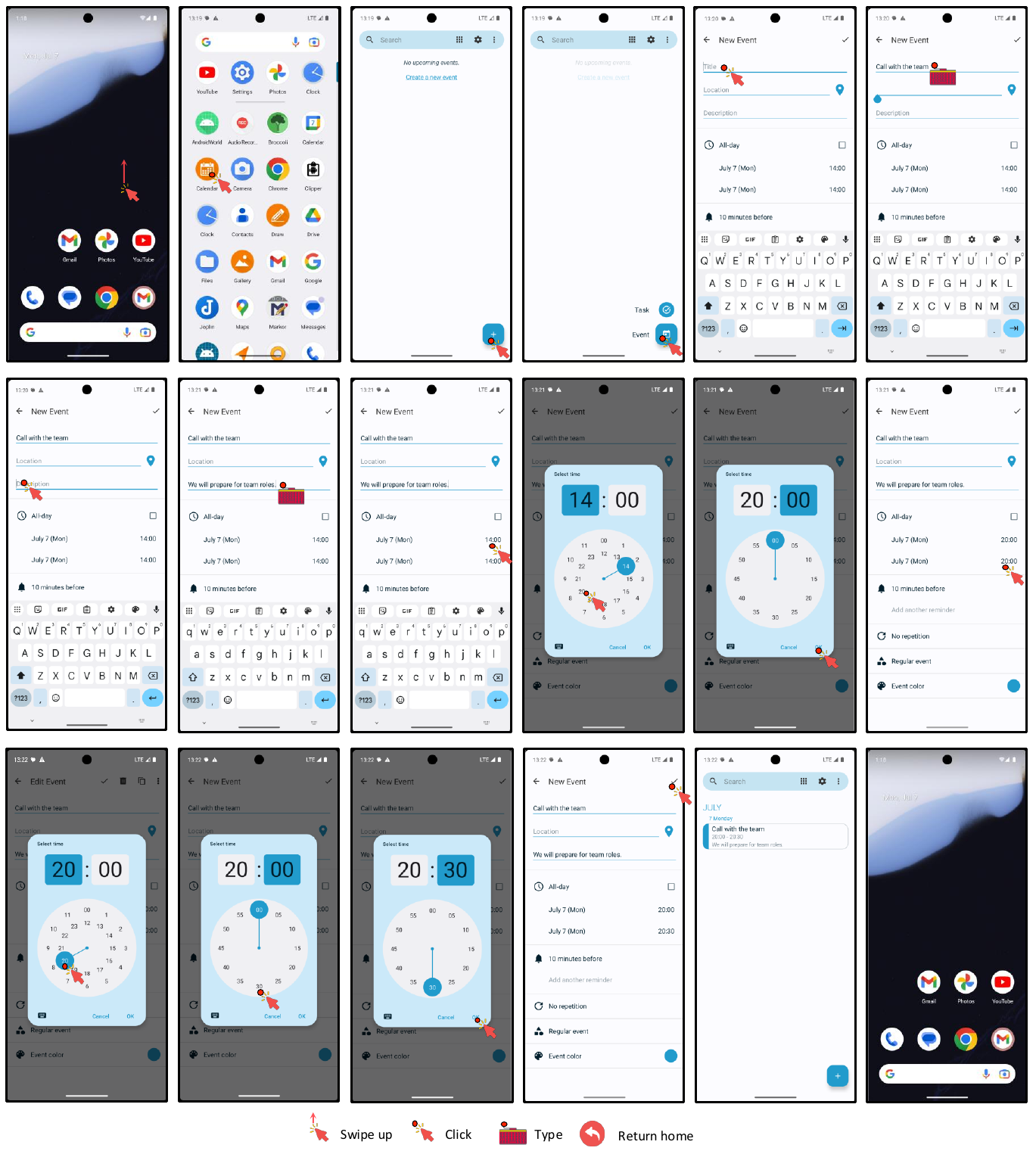}
        \label{fig:case_ours}
    \end{subfigure}
    \vspace{-0.2in}
    \caption{Case Studies. The case illustrates the task: ``Create a calendar event for tomorrow at 20h with the title 'Call with the Team' and the description 'We will prepare for team roles.'. The event should last for 30 mins.'' The left shows the execution before reinforcement learning, while the right shows the result after RL (ours). The pre-RL agent misses two critical steps: (1) omitting the meeting description, and (2) failing to set the event’s end time.}
    \label{fig:curriculum_plots}
\end{figure}

\end{document}